\documentclass[journal]{IEEEtran}
\usepackage{algorithm}
\usepackage{caption}
\usepackage{subcaption}
\usepackage{algpseudocode}
\usepackage{fancyhdr}
\usepackage{textcomp}
\usepackage{amsmath}
\usepackage{amssymb,amsfonts}
\usepackage{graphicx}
\usepackage{textcomp}
\usepackage{xcolor}
\usepackage{multicol}
\usepackage{multirow}
\usepackage{blindtext}
\usepackage{adjustbox}
\usepackage{array} 

\usepackage[maxnames=6, minnames=6,backend=bibtex,style=ieee]{biblatex}
\algnewcommand{\Initialize}[1]{%
  \State \textbf{Initialize:}
\parbox[t]{.8\linewidth}{\raggedright #1}
}
\algnewcommand{\Goto}{\textbf{go to}}%

\def\BibTeX{{\rm B\kern-.05em{\sc i\kern-.025em b}\kern-.08em
    T\kern-.1667em\lower.7ex\hbox{E}\kern-.125emX}}

\usepackage{graphicx}

\begin{document}
%
\title{Generative AI-in-the-loop: Integrating LLMs and GPTs into the Next Generation Networks}
%
%
%

    
    
    
\author{Han~Zhang,
        Akram Bin Sediq,
        Ali Afana,
        and~Melike~Erol-Kantarci,~\IEEEmembership{Senior Member, IEEE}
\thanks{Han Zhang and Melike Erol-Kantarci are with the School of Electrical Engineering and Computer Science, University of Ottawa, Ottawa, ON K1N 6N5, Canada (e-mail: hzhan363@uottawa.ca; melike.erolkantarci@uottawa.ca).}
\thanks{Akram Bin Sediq and Ali Afana are with Ericsson, Ottawa, K2K 2V6, Canada(e-mail:
akram.bin.sediq@ericsson.com; ali.afana@ericsson.com)
}}


\markboth{ }%
{Shell \MakeLowercase{\textit{et al.}}: Bare Demo of IEEEtran.cls for IEEE Journals}

\maketitle

\begin{abstract}
In recent years, machine learning (ML) techniques have created numerous opportunities for intelligent mobile networks and have accelerated the automation of network operations. However, complex network tasks may involve variables and considerations even beyond the capacity of traditional ML algorithms. On the other hand, large language models (LLMs) have recently emerged, demonstrating near-human-level performance in cognitive tasks across various fields. However, they remain prone to hallucinations and often lack common sense in basic tasks. Therefore, they are regarded as assistive tools for humans. In this work, we propose the concept of “generative AI-in-the-loop” and utilize the semantic understanding, context awareness, and reasoning abilities of LLMs to assist humans in handling complex or unforeseen situations in mobile communication networks. We believe that combining LLMs and ML models allows both to leverage their respective capabilities and achieve better results than either model alone.
To support this idea, we begin by analyzing the capabilities of LLMs and compare them with traditional ML algorithms. We then explore potential LLM-based applications in line with the requirements of next-generation networks. We further examine the integration of ML and LLMs, discussing how they can be used together in mobile networks. Unlike existing studies, our research emphasizes the fusion of LLMs with traditional ML-driven next-generation networks and serves as a comprehensive refinement of existing surveys. Finally, we provide a case study to enhance ML-based network intrusion detection with synthesized data generated by LLMs. Our case study further demonstrates the advantages of our proposed idea.

\end{abstract}

\IEEEpeerreviewmaketitle

\section{Introduction}

With the recent worldwide deployment of 5G and the ongoing standardization of 6G networks, mobile network evolution has been accelerating at a fast pace. In recent years, machine learning (ML) algorithms have emerged as effective and promising solutions in intelligent mobile networks thanks to their superior ability to process data, provide insights and take decisions. Techniques including classification, regression, clustering, and reinforcement learning (RL) have been adopted in application areas such as anomaly detection, traffic prediction, handover control, and resource allocation. 
The widespread use of ML techniques has accelerated the automation of selected tasks and laid the foundation for artificial intelligence (AI)-native mobile communication networks \cite{whitepaper}.

The introduction of ML techniques also brought some challenges. Existing ML models are typically trained based on large datasets and designed to improve capabilities on specific tasks. Therefore, the performance of the model is highly dependent on the size and quality of the training dataset.
More importantly, under complex mobile communication environments, the input variables and considerations of some tasks are often beyond the capabilities of simple ML techniques \cite{tarkoma2023ai}. A solution to this problem is to involve human interactions in the ML-based automation process, a mechanism known as "Human in the Loop (HITL)" \cite{humanintheloop}. In this way, human operators can utilize their semantic comprehension, problem-solving skills, and context awareness to deal with complicated or unexpected situations. The downside of this mechanism is that human intervention usually comes at a high cost and the quality of human decisions highly depends on the level of expertise and human reactions are slower relative to real-time decisions. 

Large language models (LLMs), especially generalized pre-trained transformers (GPTs) have recently attracted  significant attention in the industry and academia. In addition to basic language processing skills like semantic understanding or word generation, these models are also equipped with emerging abilities, such as instruction following and reasoning, as well as augmented abilities, such as interacting with others and self-improvement. Therefore, LLMs have the potential to assist humans 
with tasks that require human intelligence \cite{chiang2023can}.

Inspired by these thoughts, in this work, we propose the concept of "generative AI-in-the-loop" 
and utilize the abilities of LLMs to assist humans in handling complex or unforeseen situations in mobile communication networks. Several existing studies have investigated the application of LLMs in communication systems. For instance, \cite{tarkoma2023ai} discussed the integration of LLMs and GPTs within 6G systems and possible applications. \cite{jiang2023large} proposed an LLM-empowered multi-agent system and leveraged the interactions between agents to enhance the task-solving capabilities in the 6G network. \cite{shen2024large} discussed how to combine LLMs with edge AI and meet users' demands with LLM-empowered code generation. These works usually make discussions at a generalized level, giving broad research directions.
Differently, our work focuses more on combining LLMs and ML models as solutions to specific problems in mobile communication networks.

\begin{figure*}[t]
\centering
\includegraphics[width=6.8in]{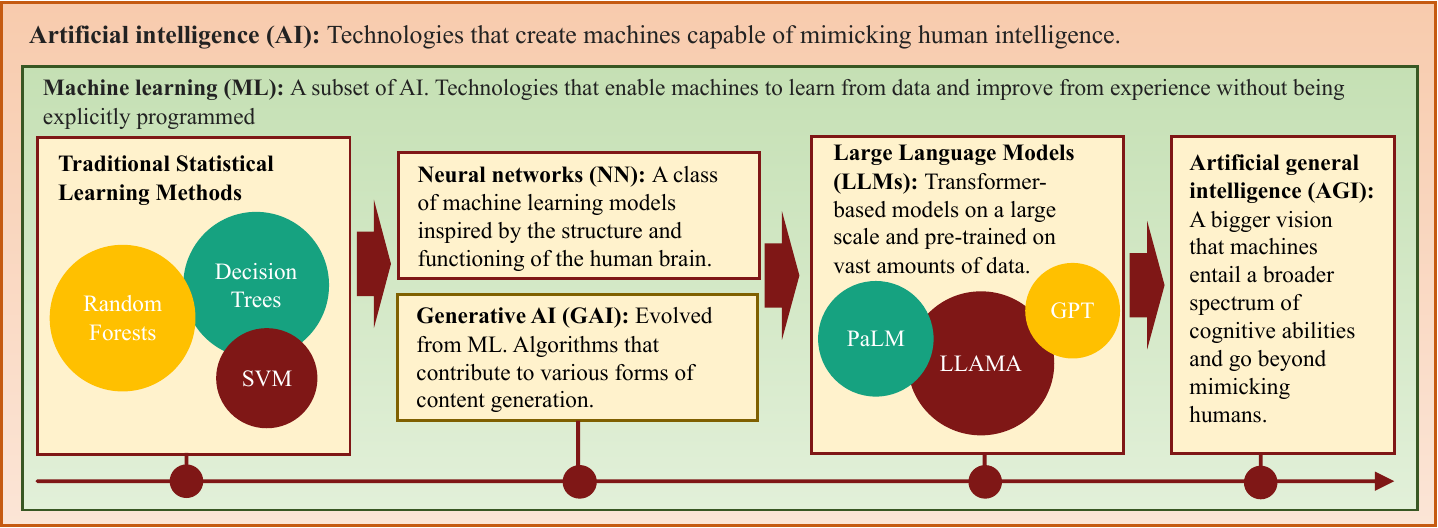}
\caption{Summary of concepts related to the topic of this work, including AI, ML, GAI, NN, LLM, GPT, and AGI.}
\label{fig0}
\end{figure*}

The main contributions of this work are summarized as follows:
\begin{itemize}
    \item We propose the concept of "generative AI-in-the-loop". This concept involves leveraging LLMs to assist humans in managing and controlling complex network tasks that are typically beyond the capabilities of traditional ML techniques. This idea accelerates the full automation of network applications and takes a big step toward the ultimate goal of AI-native mobile communication networks.
    \item Unlike existing studies, our research emphasizes the fusion of LLMs with traditional ML-driven next-generation networks. We argue that although LLMs and ML models show great capabilities, they cannot replace the roles of each other. Combining the generative and reasoning capabilities of LLM with the data analytics capabilities of ML can achieve better results than either model alone can achieve. So we begin by analyzing the capabilities of LLMs and comparing them with traditional ML algorithms. We then explore potential integration by examining the life-cycle of the ML models. Our work also serves as a comprehensive refinement of existing surveys.
    \item In addition, a case study is given to enhance ML-based network intrusion detection with synthesized data generated by LLMs. It further demonstrates the advantages of our proposed idea.
\end{itemize}

The rest of the paper is organized as follows: Section II presents the concept of "generative AI-in-the-loop" according to LLMs' capabilities and the requirements of the next-generation networks. Section III discusses possible ways to combine LLMs with traditional ML models and Section IV discusses the different deployments of LLMs and ML models. Finally, Section V presents a case study of a pre-trained LLM-enhanced network intrusion detection application and Section VI concludes the paper.

\section{Generative AI-in-the-loop}

In this section, we first explain the capabilities of LLMs and traditional ML models. Then, we discuss how to implement "generative AI-in-the-loop" and leverage LLMs and GPTs to facilitate the automation of next-generation networks.

\subsection{Background and capabilities of LLMs}

LLMs refer to a group of transformer-based large-scale statistical language models that are developed from neural language models (NLMs) and are pre-trained on massive amounts of data. With the popularity of LLMs, some comparable concepts have emerged to the public. The explanations for these concepts, including AI, ML, GAI, LLM, GPT, and AGI, and their connections are summarized in Fig. \ref{fig0}. In this work, we use the term "traditional ML techniques" to refer to the traditional statistical algorithms, reinforcement learning and others, used before the emergence of LLMs. 

\begin{table*}[t]
\centering
\begin{tabular}{|c|l|l|}
\hline
\multicolumn{1}{|l|}{} &
  \multicolumn{1}{c|}{Benefits} &
  \multicolumn{1}{c|}{Concerns} \\ \hline
\multirow{4}{*}{Pre-trained LLMs} &
  \begin{tabular}[c]{@{}l@{}}• Ability to handle unforeseen problems: LLMs can \\ handle highly dynamic environments and deal with\\  unforeseen problems.\end{tabular} &
  \begin{tabular}[c]{@{}l@{}}• Unstable results: The sampling methods make LLMs naturally\\  randomized and have high stochastic outputs\end{tabular} \\ \cline{2-3} 
 &
  \begin{tabular}[c]{@{}l@{}}• Less data dependency: Pre-trained LLMs can accomplish new\\ tasks with just a few contextual examples and do not\\  need comprehensive training processes. \end{tabular} &
  \multirow{2}{*}{\begin{tabular}[c]{@{}l@{}}• High Computation cost: The extremely large size of LLMs\\ result in a higher cost for training and fine-tuning, and limits \\ the deployments on local servers.\end{tabular}} \\ \cline{2-2}
 &
  \begin{tabular}[c]{@{}l@{}}• Semantic understanding of input variables: LLMs\\ can easily understand the meaning of different input \\ variables and their relations with the given task \\ without task-specific training.\end{tabular} &
   \\ \cline{2-3} 
 &
  \begin{tabular}[c]{@{}l@{}}• Explainable outputs and decisions: LLMs can \\ generate explanations for outputs in human-like\\  language forms, enhancing the explainability of \\ the decisions.\end{tabular} &
  \begin{tabular}[c]{@{}l@{}}• Hallucinations: In some tasks, LLMs may generate outputs\\ that are nonsensical, inaccurate, or disconnected from inputs. \\Data quality is critical to reducing hallucinations. \end{tabular} \\ \hline
\multirow{3}{*}{\begin{tabular}[c]{@{}c@{}}Traditional \\ ML models\end{tabular}} &
  \begin{tabular}[c]{@{}l@{}}• Smaller size and easier to train: Compared with LLMs, \\ traditional ML models usually have smaller sizes. That\\ makes models easier to be trained and inferred locally.\end{tabular} &
  \begin{tabular}[c]{@{}l@{}}• Critical data demands: The quality of traditional ML models \\ greatly depends on the quality of training data. ML models are \\ vulnerable to inadequate training data, data quality issues, and \\ algorithmic biases.\end{tabular} \\ \cline{2-3} 
 &
  \begin{tabular}[c]{@{}l@{}}• Deterministic output: Many of the existing ML models\\ are deterministic models and have higher reproducibility.\end{tabular} &
  \begin{tabular}[c]{@{}l@{}}• Limited capabilities: Traditional ML models have restricted \\ capabilities and can be only used to process numerical inputs.\end{tabular} \\ \cline{2-3} 
 &
  \begin{tabular}[c]{@{}l@{}}• Data analytical ability: Traditional ML models are \\ more suitable for analyzing large datasets and \\ capturing specific patterns in high-dimensional data.\end{tabular} &
  \begin{tabular}[c]{@{}l@{}}• Lack of flexibility: Traditional ML models are usually \\ trained for a single task. When the environment or task \\ objectives change, the old ML model is no longer\\ applicable and needs to be retrained.\end{tabular} \\ \hline
\end{tabular}
\caption{A comparison of the benefits and concerns using LLMs and traditional ML models }
\label{tab1}
\end{table*}

Benefiting from the remarkably large model size and enriched training data, LLMs possess a couple of critical capabilities that have not appeared in previous NLMs. This makes LLMs not only effective language processing tools but also all-purpose task-solvers. The capabilities of LLMs are concluded as follows \cite{minaee2024large}:
\begin{itemize}
    \item Understanding and generation: Language understanding and generation are the two most basic abilities of LLMs. It means that LLMs are capable of understanding human language without intentional instructions and generating outputs in a similar form. 
    Furthermore, recent tools are combining LLMs with pre-trained multi-modal foundation models. As a result, the understanding and generation capabilities of LLMs are not only limited to the text modality but also can be extended to other modalities like images, videos, and audio.
    \item Reasoning: Reasoning abilities such as planning and logical thinking enable LLMs to handle complex tasks like solving mathematical problems and target decomposition. These abilities can be promoted by skillful prompting designs including chain-of-thought (CoT), tree-of-thought (ToT), self-consistency, and in-context learning (ICL).
    \item Powerful knowledge base: The training process of LLMs can be seen as a compression and abstraction of large amounts of training text. As a result, extensive knowledge is stored in the billions of parameters in LLMs and serves as the basis for LLM-enabled problem-solving. The knowledge base can also be continuously updated and replenished with additional training data through fine-tuning.
    \item Social ability: Social ability means LLMs can communicate and interact with other LLMs or humans in an intelligible way. It is derived from a combination of understanding and generation ability, reasoning ability, and the powerful knowledge base of LLMs. 
    This capability facilitates the development of LLM-based AI agents for fully automated decision-making and control.
\end{itemize}

These above-mentioned capabilities serve as 
convincing basis for the feasibility of "generative AI-in-the-loop" in mobile networks.

\subsection{Traditional ML algorithms and LLMs}
Although LLMs show strong capabilities, the deployments are usually limited by high cost, large scale, and unstable results, which prevent LLMs from completely replacing traditional ML models in many tasks. 
To make LLMs a good-fit in ML-driven mobile networks, the benefits and concerns of using LLMs and traditional ML models are first compared in Table \ref{tab1}. 

As shown in this table, LLMs have the benefits of enhanced capabilities, less data dependency, semantic understanding of input variables and explainable outputs and decisions. Traditional ML models have the benefits of smaller model size, deterministic output and data analytical ability. This gives us an insight into how they can be combined to solve mobile network-related problems.

\begin{figure*}[t]
\centering
\includegraphics[width=6.6in]{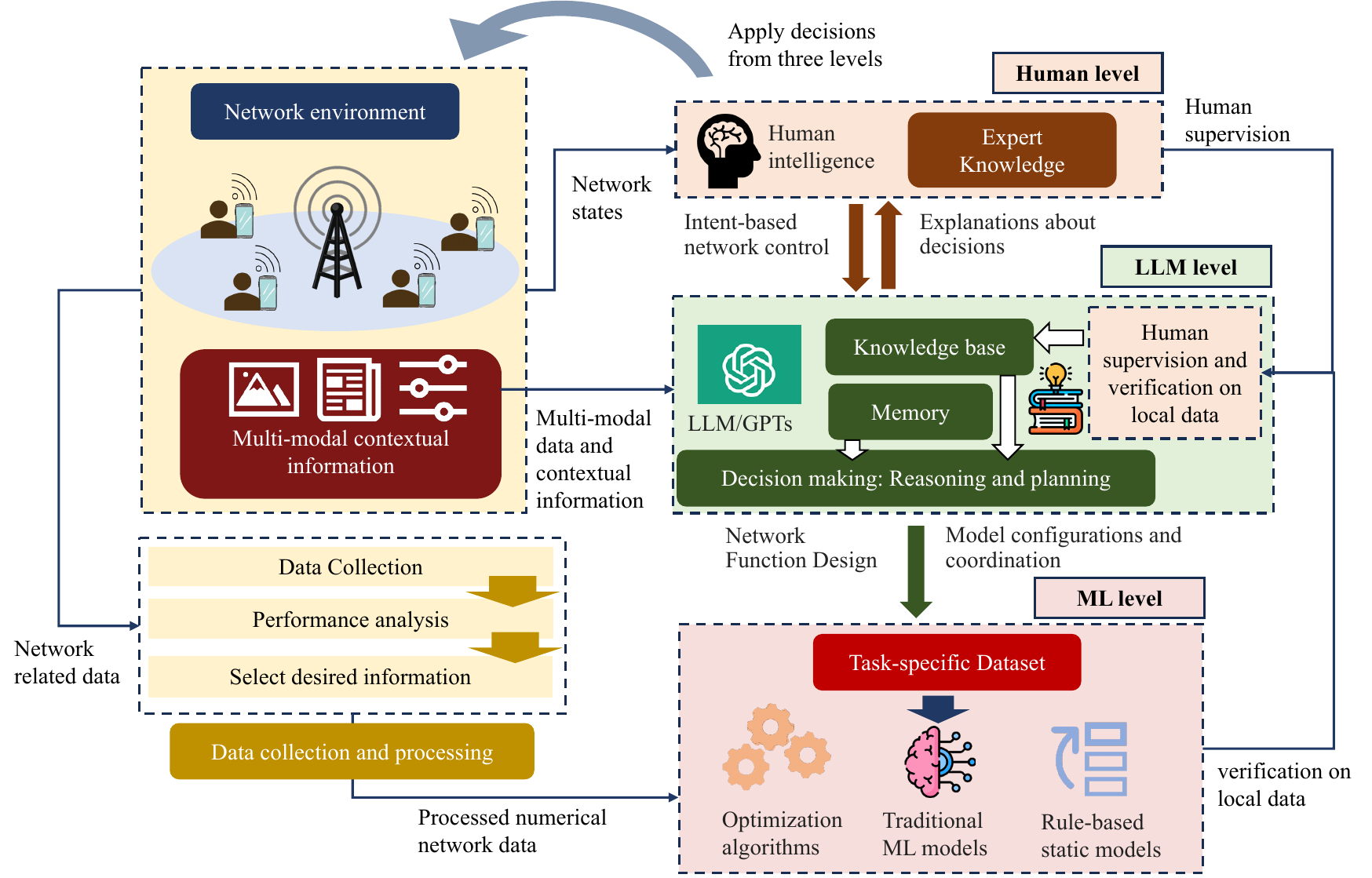}
\caption{An illustration of “generative AI-in-the-loop” in the next-generation network. LLMs act as an intermediary between human-level management and traditional ML and optimization algorithms in several ways: automate network control based on semantic intentions, generate semantic-based explanations, and perform model and network management.}
\label{fig1}
\end{figure*}

\subsection{Integrate LLMs into ML-driven next-generation network}
While ML techniques have greatly improved the efficiency of network operations, LLMs come with additional capabilities that can further accelerate the development and full automation of the next generation mobile networks.
On the other hand, the outputs of LLMs, especially GPTs, may be perceived as demonstrating intelligence that is close to human levels.
 In a few existing studies \cite{chiang2023can} \cite{suri2024large}, LLMs have demonstrated comparable performance with that achieved by human experts for some specific tasks, even though they can fail at simple tasks for humans or judgement that involves common sense. Nevertheless, in well-defined tasks, this promotes the possibility of "generative AI-in-the-loop", for instance, LLMs to assist humans for network management in slow timescales.

Fig. \ref{fig1} shows an illustration of “generative AI-in-the-loop” in the next-generation networks. As can be observed, there are three optional levels of network configurations, human level, LLM level, and ML level. All these three levels can observe desired information from the network environment and they can also interact with each other for collaborative network management. Other than the traditional ML models, the ML level may also include optimization algorithms, rule-based static models, or other small-size mathematical models. This level directly collects data from the network environment in numeric format, conducts performance analysis, and selects desired information 
for decision-making. The parameters of optimization algorithms and the rules of static models can be set manually by human experts or tuned by the LLMs. The ML models can be trained either online or offline with task-specific data. The ML level holds the advantage of short inference time and low inference cost. As a result, applications with stricter timing requirements are suitable to be deployed at this level, including beam management, radio link monitoring, and user scheduling.

On top of the ML level, the LLM level is set. The LLM level can acquire and comprehend multi-modal input from the network environment. It also provides high-level guidance for the ML level by leveraging the strong knowledge base and memory within the context window for network task-related planning, reasoning, and decision-making. For example, it can guide the design of ML models or optimization algorithms, decide the rules for rule-based static models, or perform coordination between different models. In addition, considering the uneven development and instability of most LLMs, supervisory mechanisms need to be established to verify the effectiveness of LLMs' outputs and avoid hallucinations. Possible supervisory methods include human supervision or verification of local data.

The human level is set at the very top of the framework. It is equipped with expert knowledge of mobile networks and can involve considerations related to human factors such as customer demands and commercial costs. The LLM level acts as an interface between the human level and lower levels. It can translate intent-based network control into manageable tasks for ML models and generate explanations of ML layer decisions for humans.

Within the "generative AI-in-the-loop" framework, the combination of LLMs and traditional ML models can capitalize on the benefits of both LLMs and traditional ML models and compensate for the weaknesses of the other.

\section{Leveraging LLMs for the next-generation networks}
In this section, we will first analyze some major issues in ML-based mobile network management, and then introduce potential LLM-based applications according to the requirements of next-generation networks. 

\subsection{Open issues in ML-based mobile network management}
In this subsection, some open issues in ML-based mobile network management \cite{liu2023deep} and possible LLM-based solutions for these issues are first discussed below:

1) Scarcity of high-quality training data: 
ML-based network management approaches usually require large quantity of high-quality data for training the models. However, public datasets in this field, especially labeled datasets, are very scarce. Specifically, some datasets are outdated for network management because they were collected before some major technological changes.

One possible solution is to use LLMs to augment or expand datasets for model training. LLMs can also be leveraged to identify anomaly samples in the dataset, improve the data quality, and synthesize new training data.

2) Limited Flexibility: Some traditional ML models are trained under the ideal network environment settings and they may not apply to realistic sub-optimal communication channels. 
Moreover, mobile networks are highly dynamic over time considering the mobility of devices. 

LLMs can solve these problems by generating more intelligent network management strategies. For instance, LLMs can extract constant latent representations from the dynamic network states and use them as new inputs for traditional ML models.

3) Security: Increasing diversity and complexity of mobile networks have brought new security concerns. 
Specifically, deploying ML models opens mobile networks to external data and introduces new attacks such as data poisoning attacks and membership inference attacks. This may exacerbate the security challenges for the network.

To mitigate these security concerns, on the one hand, LLM-based intelligent security monitoring schemes can be used. On the other hand, some ML-based functions can be replaced by LLMs since locally-run LLMs are better encapsulated and are less vulnerable to traditional attack methods.

\subsection{Integrate LLMs into next-generation networks}

Inspired by the above analysis of issues and solutions, there are three different ways to integrate LLMs into next-generation networks:

1) Develop LLMs-based network functions: LLMs can be directly used to perform network functions that are usually developed with traditional ML models or rule-based static models. \cite{li2023label} verifies that LLMs can be used for precise label prediction after fine-tuning. \cite{yao2022react} shows that LLMs are able to perform some simple decision-making tasks by alternating reasoning and acting. These studies reflect the ability of LLMs to carry out some network functions previously based on traditional ML models. 

In addition, LLMs can be integrated with ML models and used to enhance the performance of traditional ML models. In this way, the data scarcity and flexibility limitations of traditional ML models can be mitigated. Specific integration methods are discussed in the following sections. Potential LLM-based network functions include traffic forecasting, anomaly detection, security monitoring, network automation, and content compression.

2) LLM-assisted network application design: LLMs can also be used as effective tools while designing network applications. For instance, they can be leveraged for code generation and simulation system modeling. They can also be used to decide ML model structures, hyper-parameters, or the rules for developing static models. Additionally, LLMs 
can break down large complex tasks into small manageable tasks, and distribute tasks to different ML models \cite{tarkoma2023ai}.

3) Semantic understanding-empowered network management:
Considering the strong language understanding ability of LLMs, they can be used for semantic understanding in network management. For instance, LLMs can extract configuration information or optimization objectives from human-like input. 

4) Build general-purpose AI agents for network nodes: Another way to integrate LLMs into next-generation networks is to build general-purpose AI agents on network nodes. By deploying LLMs on different network nodes, the agents are expected to adapt to dynamic environmental changes and perform different network tasks, even without intentional and specialized training. The agents are also expected to interact with each other and improve their performance on tasks through interactions with the environment and with other agents in the absence of human intervention \cite{shen2024large}. This can lead to a agent-team-driven radio access network (RAN). 

In the following sections, we will look more deeply into the combination of ML and LLMs and discuss how they can be simultaneously deployed in mobile networks.

\begin{figure*}[t]
\centering
\includegraphics[width=6.8in]{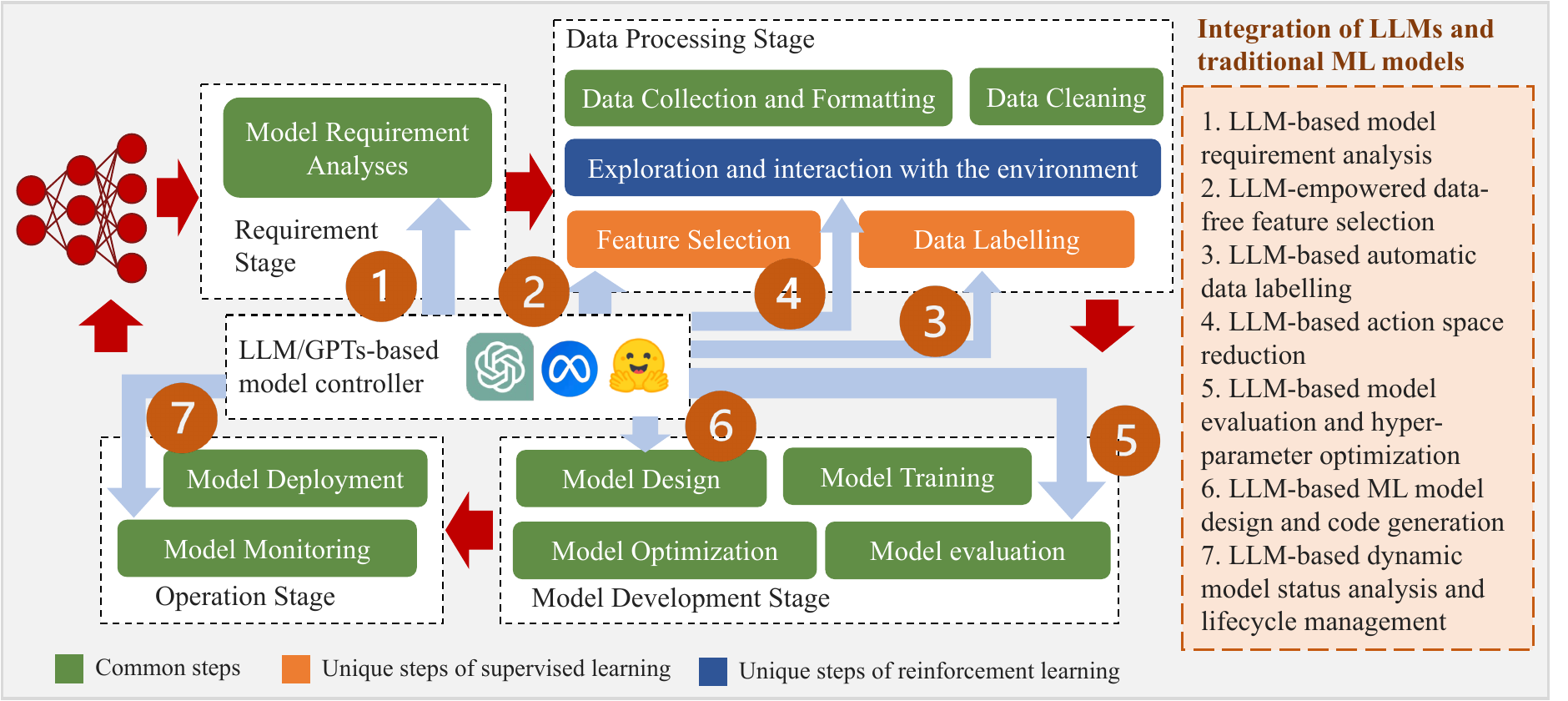}
\caption{Different ways to enhance ML models with LLMs. The life cycle of ML models includes four stages: requirement stage, data processing stage, operation stage, and model development stage. LLMs can be integrated into each stage.}
\label{fig3}
\end{figure*}

\section{Enhance traditional ML models with LLMs}
In this section, we discuss the implementation of "generative AI-in-the-loop" from the perspective of traditional ML techniques. We illustrate how LLMs can be efficiently integrated into traditional ML model design. 

Fig. \ref{fig3} shows seven different ways to enhance ML models with LLMs. LLMs can use one of these ways or a combination of all of these ways to play a supporting role to assist. 
In this figure, we divide the life cycle of traditional ML models into four stages: the requirement stage, the data processing stage, the operation stage, and the model development stage. For different machine learning methods, such as supervised learning, unsupervised learning, and RL, the steps at the data processing stage can be different. In figure, we label the steps of different learning methods with different colors. LLMs can play different roles at each stage, which is explained as follows.

\subsection{Requirement stage}
In the requirement stage, the model requirement analysis is performed to clarify the task requirements according to the task description. This helps decide the number and type of ML models needed for the given task. 
These designs are traditionally manually done with the help of human experts. However, with the emergence of LLMs, LLM-based model requirement analysis can be performed to decompose semantic-based task descriptions into several small, manageable tasks to assist humans. LLMs can also select suitable models for each task. With the integration, LLMs take on a part of the model requirement analysis work that would have been manually performed. We refer this as "generative AI-in-the-loop" to accelerate the automation of mobile networks.

\subsection{Data processing stage}
In the data processing stage, two common steps are performed. In the first step, data is collected from mobile networks and standardized. Next, data cleaning is conducted to remove incomplete or anomalous samples from the data set. LLMs can help with data cleaning by evaluating the plausibility of the collected data samples. 

In addition, there are two unique steps of supervised learning models, feature selection and data labeling.
In traditional workflows, feature selection is usually performed by analyzing the correlation between features and the output. However, this may lead to over-fitting problems and it is challenging to decide the statistical measures \cite{brownlee2019choose}. Instead, LLMs can be used for data-free feature selection based on the semantic understanding of given features. 
Data labeling is to add informative labels to the raw data so that ML models can learn from it. The traditional data labeling approach is to ask humans to recognize unlabeled data, which results in low scalability and high expenses. In comparison, LLM-based automatic data annotation, or hybrid labeling with both human experts and LLMs can automate the labeling process and lower the cost \cite{wang2021want}. 

Apart from this, RL models usually collect data by taking actions and receiving feedback from the environment. As a result, an exploration and interaction with the environment step is performed during this stage. In this step, LLMs can help with action space reduction and increase the efficiency of the exploration. Furthermore, LLMs can also help synthesize training data when there is a shortage of available datasets. 

\subsection{Model development stage}
The third stage is the model development stage. It includes model design, training, optimization, and evaluation. In the first two steps, the architecture of the ML model is decided and the model is trained with the data prepared in the previous stage. LLMs can be applied to model design through code generation. After that, the ML model is evaluated and optimized. LLM-based model evaluation can help choose suitable evaluation metrics, analyze the performance and automatically tune the hyper-parameters.

\begin{figure*}[t]
\centering
\includegraphics[width=6.8in]{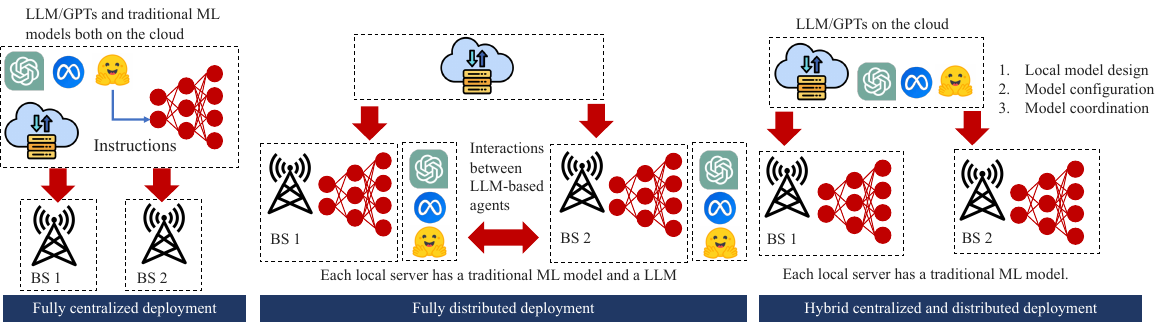}
\caption{Three ways to combine LLMs with ML models. (a) Both LLMs and ML models are deployed at the center. They are combined for network management. (b) Both LLMs and ML models are deployed in a distributed manner for Multi-Agent interaction. (c) LLMs are deployed at the center for management, and ML models are deployed in a distributed manner for local training and inference.}
\label{fig4}
\end{figure*}

\begin{figure*}[]
\centering
 \begin{subfigure}[b]{0.45\textwidth}
     \centering
     \includegraphics[width=\textwidth]{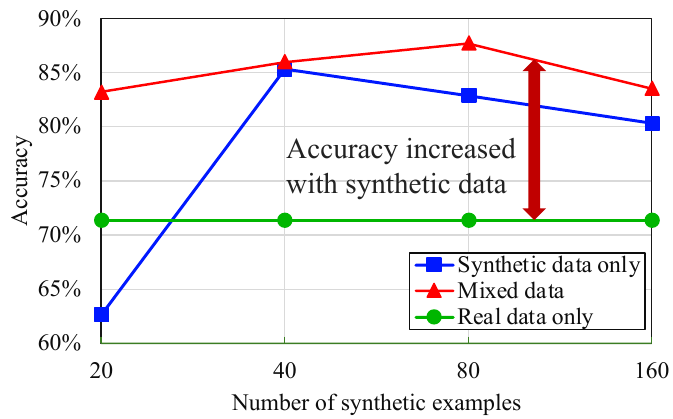}
     \captionsetup{font={small}}
     \captionsetup{font=scriptsize}
     \caption{Accuracy of network intrusion detection under different numbers of synthetic examples.}
     \label{fig6-4}
 \end{subfigure}
  \hfill
 \begin{subfigure}[b]{0.45\textwidth}
     \centering
     \includegraphics[width=\textwidth]{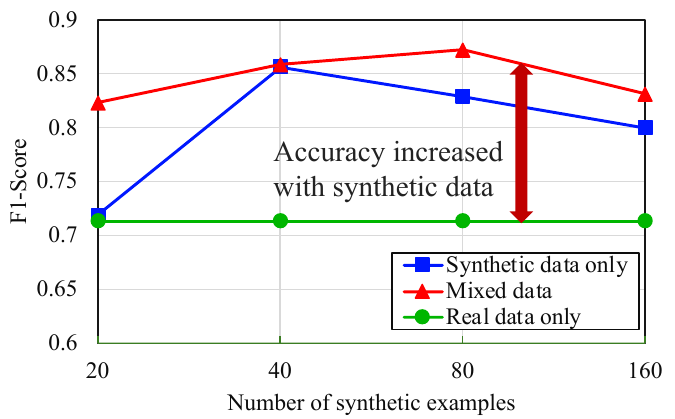}
     \captionsetup{font=scriptsize}
     \caption{F1-score of network intrusion detection under different numbers of synthetic examples.}
     \label{fig6-5}
 \end{subfigure}

 \begin{subfigure}[b]{0.45\textwidth}
     \centering
     \includegraphics[width=\textwidth]{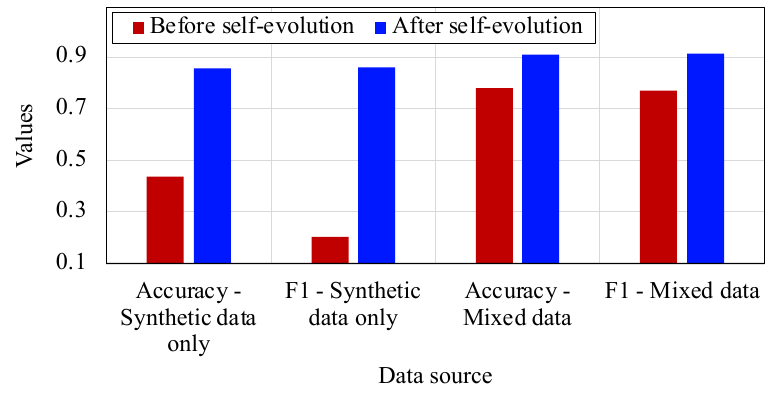}
     \captionsetup{font={small}}
     \captionsetup{font=scriptsize}
     \caption{Accuracy and F1-score of network intrusion detection using the synthetic data before self-evolution and after self-evolution.}
     \label{fig6-1}
 \end{subfigure}
\hfill
 \begin{subfigure}[b]{0.45\textwidth}
     \centering
     \includegraphics[width=\textwidth]{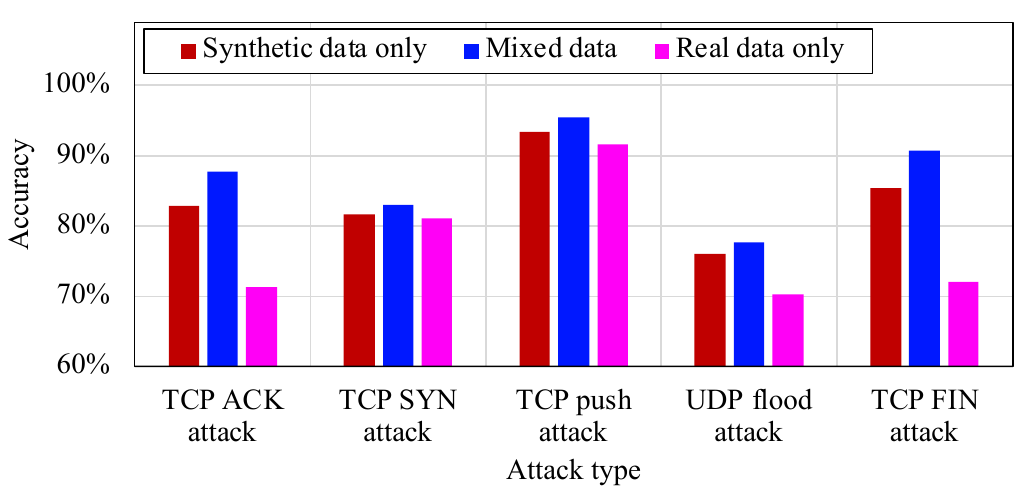}
     \captionsetup{font={small}}
     \captionsetup{font=scriptsize}
     \caption{Accuracy of network intrusion detection using synthetic data generated by GPT-3.5 under different types of attacks.}
     \label{fig6-2}
 \end{subfigure}

 \begin{subfigure}[b]{0.45\textwidth}
     \centering
     \includegraphics[width=\textwidth]{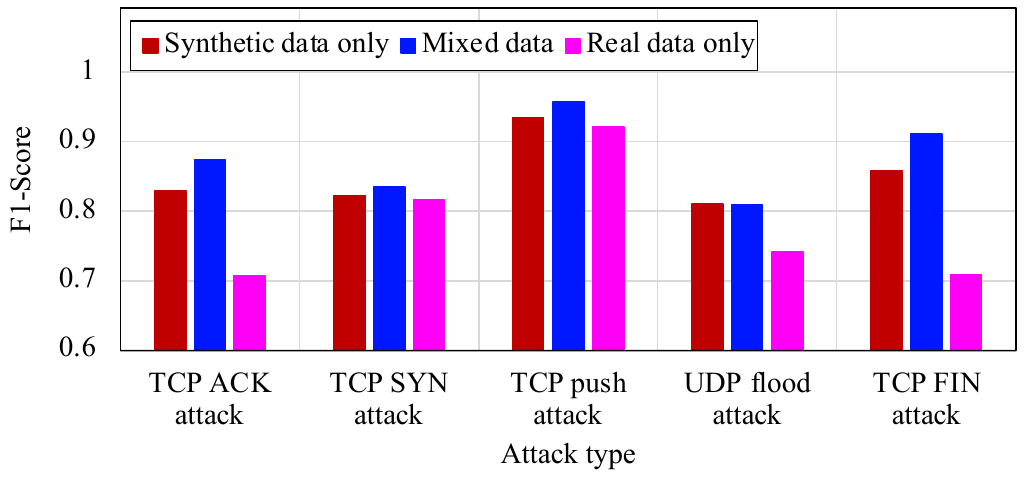}
     \captionsetup{font=scriptsize}
     \caption{F1-score of network intrusion detection using synthetic data generated by GPT-3.5 under different types of attacks.}
     \label{fig6-3}
 \end{subfigure}
 \captionsetup{font={small}}
\caption{Accuracy and F1-score of network intrusion detection using synthetic data.}
\label{fig7}
\end{figure*}

\subsection{Operation stage}
The last stage is the operation stage. In this stage, the ML model is deployed at network nodes for model inference. It also needs to be monitored in case of model decay, unexpected data and attacks. In this stage, LLMs can be used for dynamic model status analysis and life-cycle management. More specifically, LLMs decide which ML model should be used depending on different cases. Model size, inference time, and computation cost can be included as considerations for model management. LLMs can also be used to analyze the ML model performance and evaluate if the model continues to function as expected in realistic mobile network scenarios. If necessary, it will initiate the life cycle of new ML models.

It is worth noting that non-ML models, such as optimization algorithms and rule-based static models can also be included in such integration. For these models, the data processing stage and the model development stage are not required, but they can be developed on demand in the requirement stage and be chosen by LLMs to perform tasks in the operation stage.

\section{Deployments of ML models and LLMs in mobile networks}

In this section, we discuss how to simultaneously deploy traditional ML models and LLMs in mobile networks. As shown in Fig. \ref{fig4}, there are three different deployment options: fully centralized deployment, mixed centralized and distributed deployment, and fully distributed deployment.

Fully centralized deployment means deploying both LLMs and traditional ML models on the cloud side and combining them for network management. The benefit is that the powerful computing and storage resources of the cloud server can support the training and deployment of both LLMs and ML models. Moreover, centralized models have access to global information and can make more intelligent decisions.
However, the limitation is that the centralized structure lacks flexibility and scalability and is not suitable to handle a large number of local servers, for example, ultra-dense networks (UDNs). Also, in this deployment, the cloud server will access sensitive data generated by base stations (BSs) and bring privacy concerns. 

This deployment method is applicable when there is a small number of BSs, and the controller wants to have centralized control of the whole network. It also fits the cases when the distributed BSs only have limited computation and storage resources. Some possible applications include interference-aware radio resource management and power control, and network slice provision.

Fully distributed deployment means deploying both LLMs and traditional ML models on the local or edge site and each local BS owns a LLM. The benefit is that the LLMs and ML models are deployed closer to the data source. The sensitive data can be kept locally to enhance data privacy. In addition, fully distributed deployment has better scalability since the workload is jointly undertaken by distributed BSs. The interactions between BSs are also promoted through multi-agent conversations \cite{wu2023autogen}. 
However, the fully distributed deployment has a much higher requirement on the resources held by the local servers. The servers should be equipped with powerful computing and storage resources to support both LLMs and ML models. Additionally, the distributed structure will introduce parallelization and synchronization steps, and expose models to more vulnerabilities.

In conclusion, this deployment method applies to scenarios with powerful local edge servers and where interactions are needed between BSs. Some suitable applications include traffic prediction and user association.

Hybrid centralized and distributed deployment combines the previous two deployment methods. In this case, LLMs are deployed on the cloud while traditional ML models are deployed on local servers. This method incorporates the benefits of both fully centralized deployment and fully distributed deployment. The cloud server is usually equipped with powerful resources to support the needs of LLMs. The distributed ML models are supported by local devices and are customized for task-oriented applications. As a result, resources can be rationally exploited and data privacy can be protected. The only concern is that the communication between local ML models and the cloud-based LLMs will result in extra communication overhead.

With this deployment, LLMs can be used for distributed ML model management. They can analyze user requirements, plan tasks, and leverage ML models for execution. 
For example, LLMs can select the appropriate scheduling algorithm models and perform conflict management between different network applications. 

\section{A case study}
This section presents a case study on enhancing network intrusion detection using synthetic data generated by LLMs. We consider the dataset proposed in \cite{farzaneh2023dtl} which is collected from a cloud-based mobile communication system. In this system, a malicious attacker can perform distributed denial-of-service (DDoS) attacks and generate malicious traffic. With the labeled data, we train a convolutional neural network (CNN) to perform network intrusion detection and decide the presence of attacks in the network \cite{farzaneh2023dtl}.

Considering the scarcity of labeled data for the given task, we assume that there are only 20 available traffic data examples. Among these examples, 10 examples are collected from benign network traffic and 10 examples are collected from malicious network traffic. To improve the intrusion detection accuracy, additional traffic data examples are synthesized by GPT-3.5 and used for CNN training. The prompt for network traffic generation consists of four parts: task description, examples listing, data explanation, and output formatting. In the first part, the task for GPT-3.5 is described as "generating some data to train an ML model for network intrusion detection". Next, available training data is listed as examples for data generation. In the third part, an explanation of the meaning of each value in the dataset is given to GPT-3.5 and the semantic information can assist in data generation. Finally, output formatting is given to limit the amount of generated data and make the generated data more manageable.

In our simulations, we first explore the impact of the amount of synthetic data on the performance. Fig. \ref{fig6-4} and \ref{fig6-5} show the performance of TCP ACK attack detection while using GPT-3.5 to generate different numbers of synthetic examples. We compare the accuracy and the F1-score while only using the real data, only using the synthetic data, and using mixed data of both synthetic data and real data. As it can be observed, the synthetic data can help improve the detection accuracy from 71.3\% to above 80\%. Meanwhile, an increase in the number of synthesized examples does not necessarily improve detection performance. This reveals that the generation ability of LLMs is limited and the quality of synthetic data may decrease as more samples are generated. 

On the other hand, we notice that synthetic data generated by GPT-3.5 sometimes has instability issues. To further ensure high-quality data generation, a self-evolution can be added by using prompts like "These examples are not accurate enough to train ML models. Can you generate better data". 

Fig. \ref{fig6-1} shows a comparison of the synthetic data before and after self-evolution for the TCP FIN attack. As can be observed, before self-evolution, the CNN model trained with synthetic data shows low accuracy and F1-score. It implies that the quality of the generated data is far inferior to the original data. In contrast, after self-evolution, the data quality is improved. It is worth noting that self-evolution does not always promote performance. When GPT-3.5 has already output high-quality synthetic data, it may repeat the previously generated data. In a few cases, it also outputs invalid and confusing data. Considering the current limited capabilities of GPT-3.5, the quality of synthesized data should be verified by the real dataset before being used for CNN training.

Finally, Fig. \ref{fig6-2} and Fig. \ref{fig6-3} show the accuracy and F1-score of network intrusion detection using
synthetic data generated by GPT-3.5 under different types of attacks while 80 synthetic examples are generated. 
For the TCP ACK attack, the accuracy and F1-Score are improved by 22.1\%, and for the TCP FIN attack, the accuracy and F1-score are improved by 28.7\%. In addition, synthetic data enhances the detection of different attacks differently. 
This is because the quality of the real data, the difficulty of the task, and the ability of LLMs to recognize attack patterns in the given examples vary for different attacks. In conclusion, the simulation results demonstrate that synthesizing data with LLMs for ML model training is an effective way to solve data scarcity issues and enhance model performance for mobile communication tasks.

\section{Conclusion}

ML techniques offer numerous opportunities for intelligent mobile networks and accelerated automation. On the other hand, LLMs and GPTs have recently garnered significant attention due to their outstanding performance in cognitive tasks across various fields. Inspired by these thoughts, we propose that combining the strengths of both LLMs and ML models can yield a synergistic effect, making the whole greater than the sum of its parts. In this work, we discuss the strengths of ML models and LLMs in different tasks and explore how to effectively combine them to address mobile network-related challenges. We also provide a case study to demonstrate the advantages of our proposed idea which utilizes generation abilities of LLMs to improve the accuracy of ML-based intrusion detection. In the future, we plan to integrate LLMs into more complex and advanced ML-driven network applications.

\section*{Acknowledgment}
This work has been supported by MITACS and Ericsson
Canada, and NSERC Canada Research Chairs program.

\begin{refcontext}[sorting = none]
\small
\printbibliography
\end{refcontext}

\end{document}